# Memotion 3: Dataset on Sentiment and Emotion Analysis of codemixed Hindi-English Memes

Shreyash Mishra*[1], S Suryavardan*[1], Parth Patwa[2], Megha Chakraborty[3], Anku Rani[3], Aishwarya Reganti[4], Aman Chadha†[5,6], Amitava Das[3], Amit Sheth[3], Manoj Chinnakotla[7], Asif Ekbal[8] and Srijan Kumar[9]

[1]*IIIT Sri City, India*
[2]*UCLA, USA*
[3]*University of South Carolina, USA*
[4]*Carnegie Mellon University, USA*
[5]*Stanford University, USA*
[6]*Amazon AI, USA*
[7]*Microsoft, USA*
[8]*IIT Patna, India*
[9]*Georgia Tech, USA*

Abstract

Memes are the new-age conveyance mechanism for humor on social media sites. Memes often include an image and some text. Memes can be used to promote disinformation or hatred, thus it is crucial to investigate in details. We introduce Memotion 3, a new dataset with 10,000 annotated memes. Unlike other prevalent datasets in the domain, including prior iterations of Memotion, Memotion 3 introduces Hindi-English Codemixed memes while prior works in the area were limited to only the English memes. We describe the Memotion task, the data collection and the dataset creation methodologies. We also provide a baseline for the task. The baseline code and dataset will be made available at https://github.com/Shreyashm16/Memotion-3.0.

.

## 1. Introduction

With the rise of social media platforms as a conduit for users to communicate their thoughts and interact with one another, the amount of hate online has also parallely proliferated. The power of free uncensored speech however can cause considerable angst in the online community by demeaning other people. A popular form of producing such harmful content is the creation

*Equal contribution.

†Work does not relate to position at Amazon.

shreyash.m19@iiits.in (S. Mishra*); suryavardan.s19@iiits.in (S. Suryavardan*); amitava@mailbox.sc.edu (A. Das)

of memes. Memes generally consist of popular images and texts associated with them that intend to spark humor among the readers. A popular definition of memes, now widely used in the field, describes them as “a group of texts with shared characteristics, with a shared core of content, form, and stance”. Broadly, “content” refers to ideas and ideologies, “form” refers to our sensory experiences such as audio or visual, and “stance” refers to the tone or style, structures for participation, and communicative functions of the meme [1]. The artistic use of images and text makes the content relatable and viral. Although initially used for comic purposes only, memes have quickly evolved as a mechanism used to taunt and demean certain sections of the society. They are also used to spread misinformation and fake news. Memes are a language in themselves, with a capacity to transcend cultures and construct collective identities between people. These shareable visual jokes can also be powerful tools for self-expression, connection, social influence and even political subversion [2].

Social media platforms have many initiatives to moderate this kind of content, but memes have managed to hold their relevance despite these efforts. Detecting hate-speech and aggression on social media platforms is a popular research field, both in academia and industry, however, memes are continuously evolving and outpace contemporary hate-classification systems because (i) they can be multi-modal in nature, (ii) they might not use explicit hate content/words but more subtler forms of aggression like satire or sarcasm, and (iii) they can contain code-mixed content (languages like Hindi, Telugu, etc. written in Latin script) which is harder to parse and detect. Code-mixed content is especially prevalent in multilingual societies.

The previous iterations of Memotion each curated 10k multi-modal memes from various social media websites like Reddit, Facebook, Imgur, and Instagram, and proposed emotion and sentiment classification tasks on these datasets. In the current iteration- Memotion 3, we add an additional layer of complexity by introducing memes that are Hindi-English code-mixed. This addition ensures that models will see data that is more current and prevalent on social media and hence improve robustness. The rest of the paper is organised as follows: we describe the related work and the task in Section 2 and Section 3, respectively. Section 4 contains the details of the dataset we collected for memotion analysis: Memotion 3; followed by a brief description of baseline models 5 and their results in Section 6. We conclude with the mention of future work and limitations, in Section 7.

## 2. Related Work

Analysis of data to extract the sentiment and emotion has gained a lot of traction in recent years. This has been majorly focused on the large amount of data generated every second, thanks to social media. Most research in this area are focused on textual modality with some inclusion of multi-lingual data aimed at determining the polarity of the given data.

Many of the existing sentiment analysis datasets are of textual in nature and are in English [3, 4] with negative/positive or neutral categories. This also includes the works on hate speech detection in English from platforms such as Twitter [5, 6, 7] that classifies tweets based on the detected racism, sexism etc. Further works in this area shed light on multilingual or code-mixed data with inclusion of languages, such as Hindi [8], Hindi-English [9, 10, 11], Spanish-English [9, 12], Malayalam-English [13] and more [14, 15, 16, 17]. Recent approaches to solve this problem mostly involve the use of deep learning [18, 19, 20] and large language models [21, 22, 23, 24].

However, social media is a multi-modal platform, as a result a combination of textual and visual data is vital to capture the context and analyse the data. Text-image pairs can be used for image captioning, sentiment analysis, hate speech detection and mitigate cyberbullying as shown by existing research [25, 26, 27, 28, 29]. Moreover, research has also been done towards sentiment and emotion analysis of video based data [30, 31].

One of the most commonly occurring formats of multi-modal data in social media is a meme. Although there has only been limited work, specifically towards memes, research in this area has grown in recent times. MultiOFF [32] is binary classification dataset that aims to detect whether memes are offensive or not. The Hateful memes dataset from Facebook [33] provides memes collected from USA based social media groups along with some manually reconstructed memes annotated for both uni-modal and multi-modal hate speech. The previous iterations of Memotion i.e. Memotion 1 [34] and Memotion 2[35, 36] drew attention to analysis of English memes that covered several categories, such as hatefulness, motivation, humour, sarcasm and overall sentiment. TamilMemes [37] is a also meme classification dataset that categorizes memes as being trolls or not, however this is one of the few datasets not in English. Some approaches toward this include [38, 39, 40, 41].

With Memotion 3, we present the first code-mixed Hinglish (Hindi-English) meme analysis dataset with 10k memes annotated for the aforementioned categories in the previous iterations of Memotion.

## 3. Memotion 3 task

A dataset of 10,000 annotated Hindi-english memes is made available. Each data point has a label for each sub-task as well as an accompanying image and text. Similar to Memotion 1 [34] and Memotion 2 [35], we consider sentiment, emotions, and their intensities. Unlike previous works, however, this iteration of the Memotion challenge focuses on Hinglish language memes. Our subtasks are as follows:

- **Task A: Sentiment Analysis** - Classify a meme as positive, negative, or neutral. Figure 1 explains the potential negative connotations of a specific meme.
- **Task B: Emotion Classification** - Classify a meme into humorous, sarcastic, offensive, or inspirational. More than one category can apply to a meme. Tasks B can be clearly understood by looking at the meme in Figure 2.
- **Task C: Scales/Intensity of Emotion Classes** - Calculate the degree to which a given emotion is being conveyed is the third task. The intensity of each emotion is shown in Figure 2.

## 4. Data

In this section we describe the data collection, annotation and data analysis.

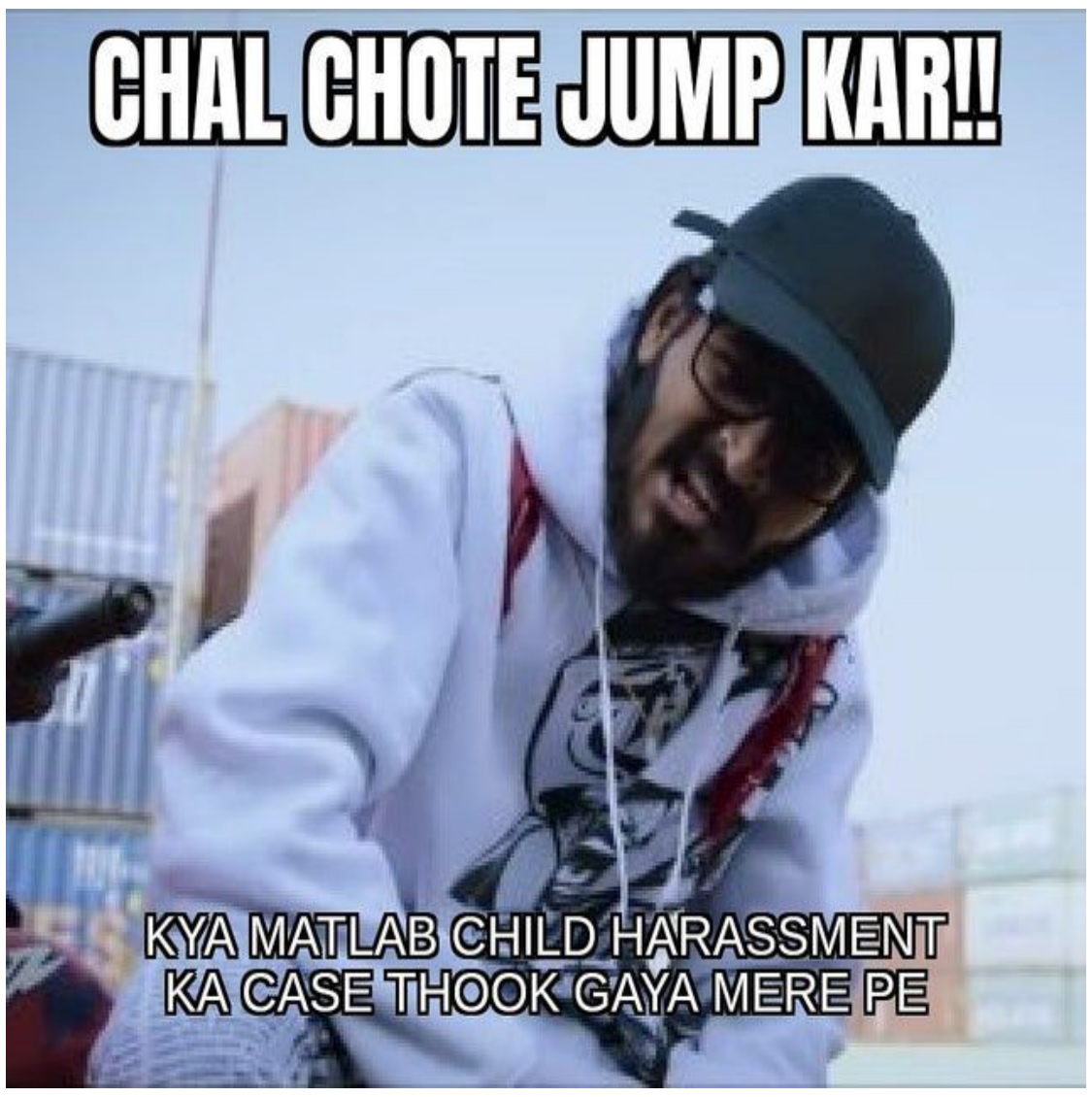

**Figure 1:** Example for Task A. People found this meme to have a negative sentiment.

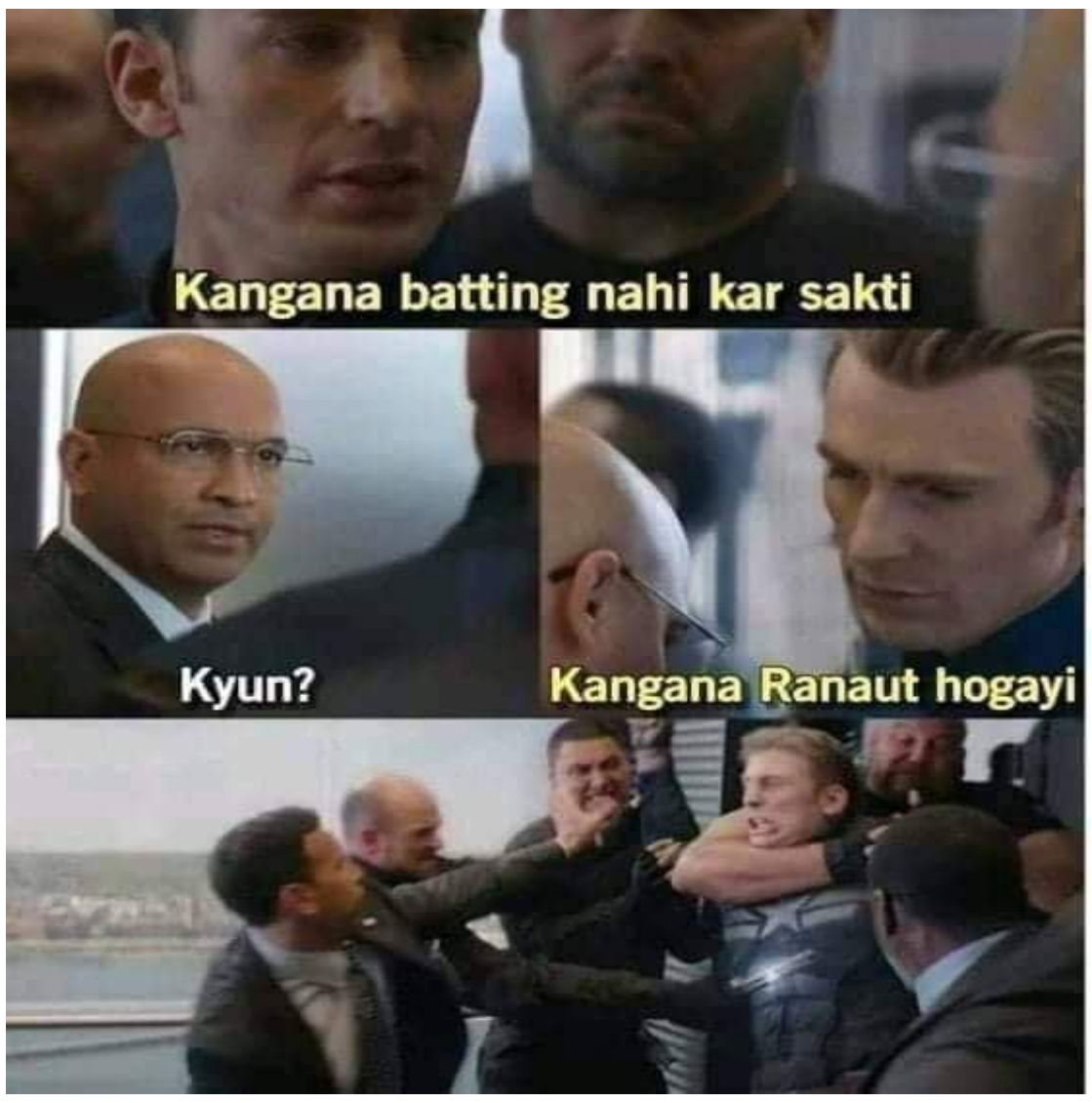

**Figure 2:** Example for Task B and C. Majority of annotators found this meme's humour intensity as very funny, sarcasm as twisted meaning, offensive as not offensive and motivational as not motivational. The corresponding labels for Task B will be funny, sarcastic, not offensive and not motivational.

## 4.1. Data Collection

We downloaded the memes after on topics of interest, such as politics, sports etc. We also collected memes using a Selenium-based web crawler. All memes are gathered from public websites Reddit and Google images. We cleaned the data to remove redundancies and performed random manual quality check. The memes are release along with the source URLs and OCR text. For OCR, we utilised the Google Vision API[1].

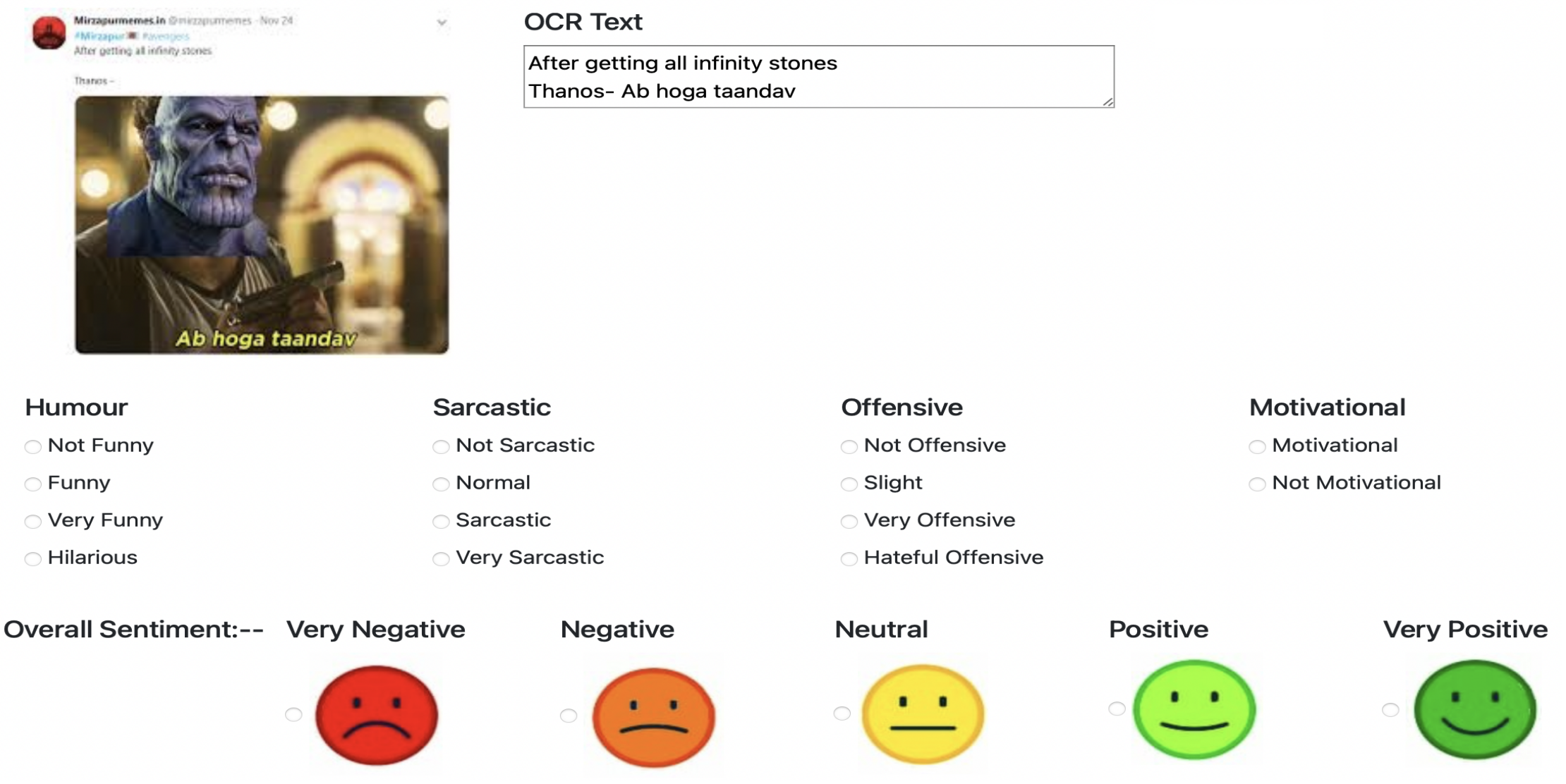

**Figure 3:** Annotator Interface. The annotators see a meme and have to mark the sentiment and emotion intensities of the meme. They also have to correct the OCR extracted using the Google Vision API, if there are any discrepancies.

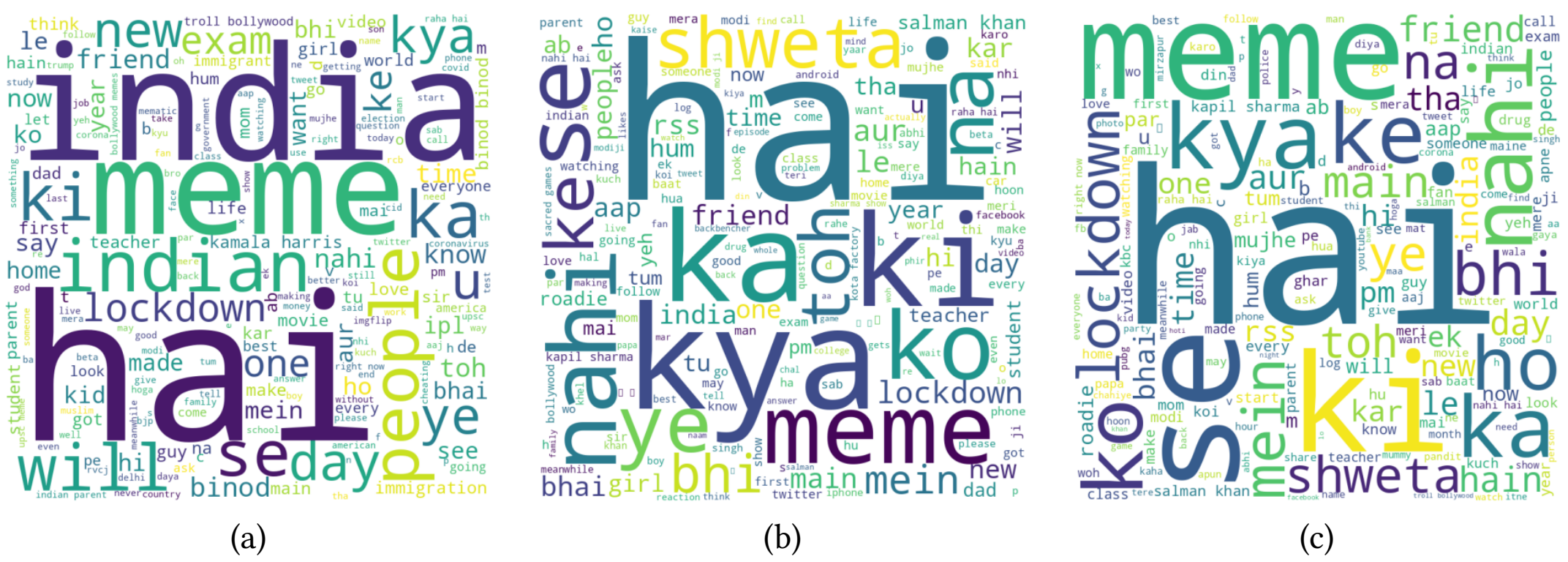

**Figure 4:** Word clouds indicating top words used for the (a) train, (b) validation and (c) test sets.

## 4.2. Data Annotation

we recruited Undergraduate student proficient in English, Hindi and meme knowledge. For annotation, they use an interface built by us, as shown in Figure 3. The annotators were asked to assess whether the meme's creator intended it to be positive, negative, or neutral in Task A.

[1]https://cloud.google.com/vision

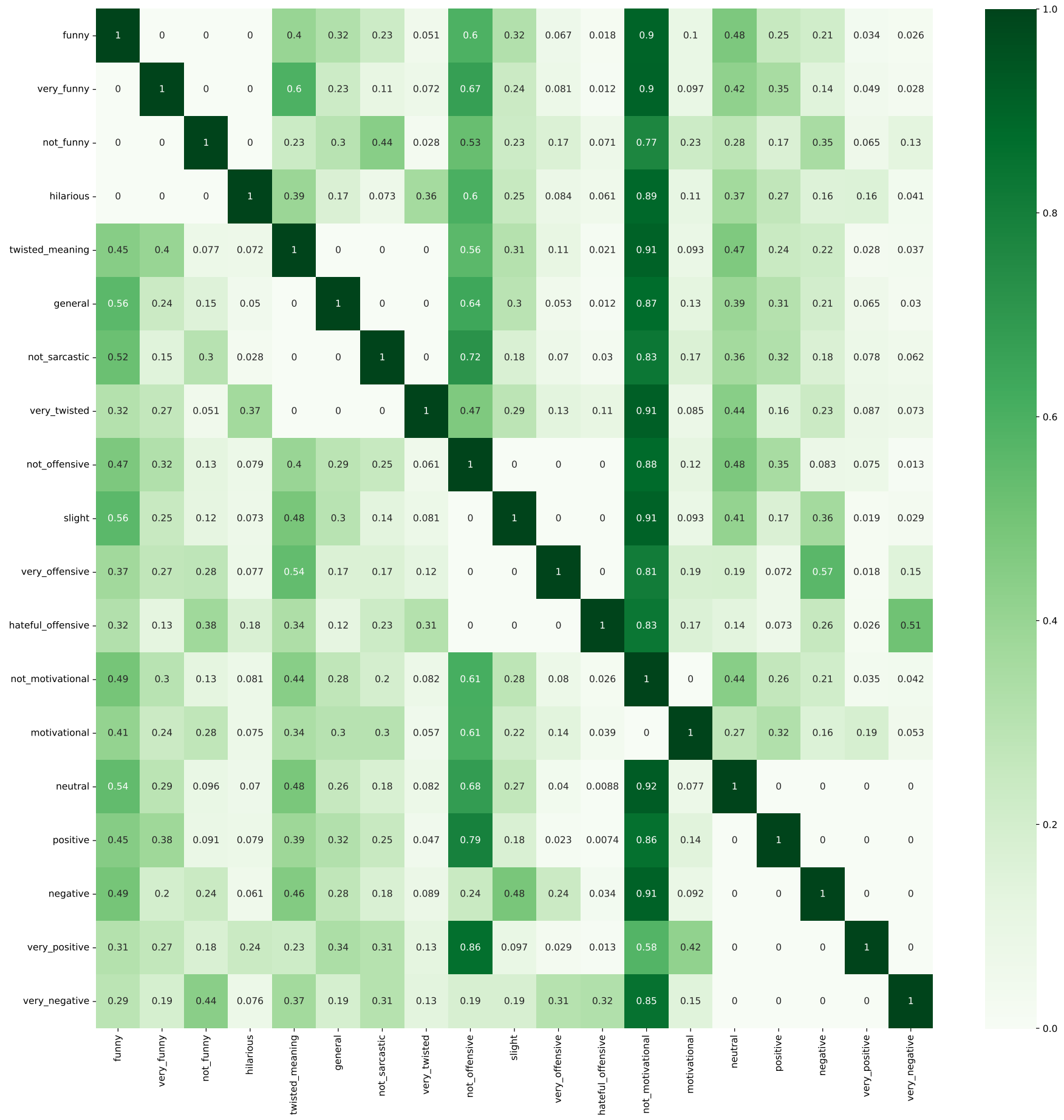

**Figure 5:** Overall distribution of the dataset showing overlap between all 20 labels

The annotators were asked to provide their thoughts on the emotion of the meme for Tasks B and C. The perception of a meme and societal elements may vary from person to person. Each meme is annotated by three separate annotators. The decision of the final annotations is made using a majority voting system.

### 4.3. Data Distribution and Analysis

The dataset consists of 10,000 memes, which are split into train, validation and test sets of size 8500,1500 and 1500 respectively. Annotations for each meme include its overall sentiment (positive, neutral, or negative), emotion (humour, sarcasm, offence, or motivation), and scale of emotion (0-4 levels). Fig. 6 shows the distribution of memes across all the labels. Fig. 4 displays the word occurrence in the dataset. From the wordcloud we can see that lot of code mixed words like *nahi*, *kya* are prominent in the dataset.

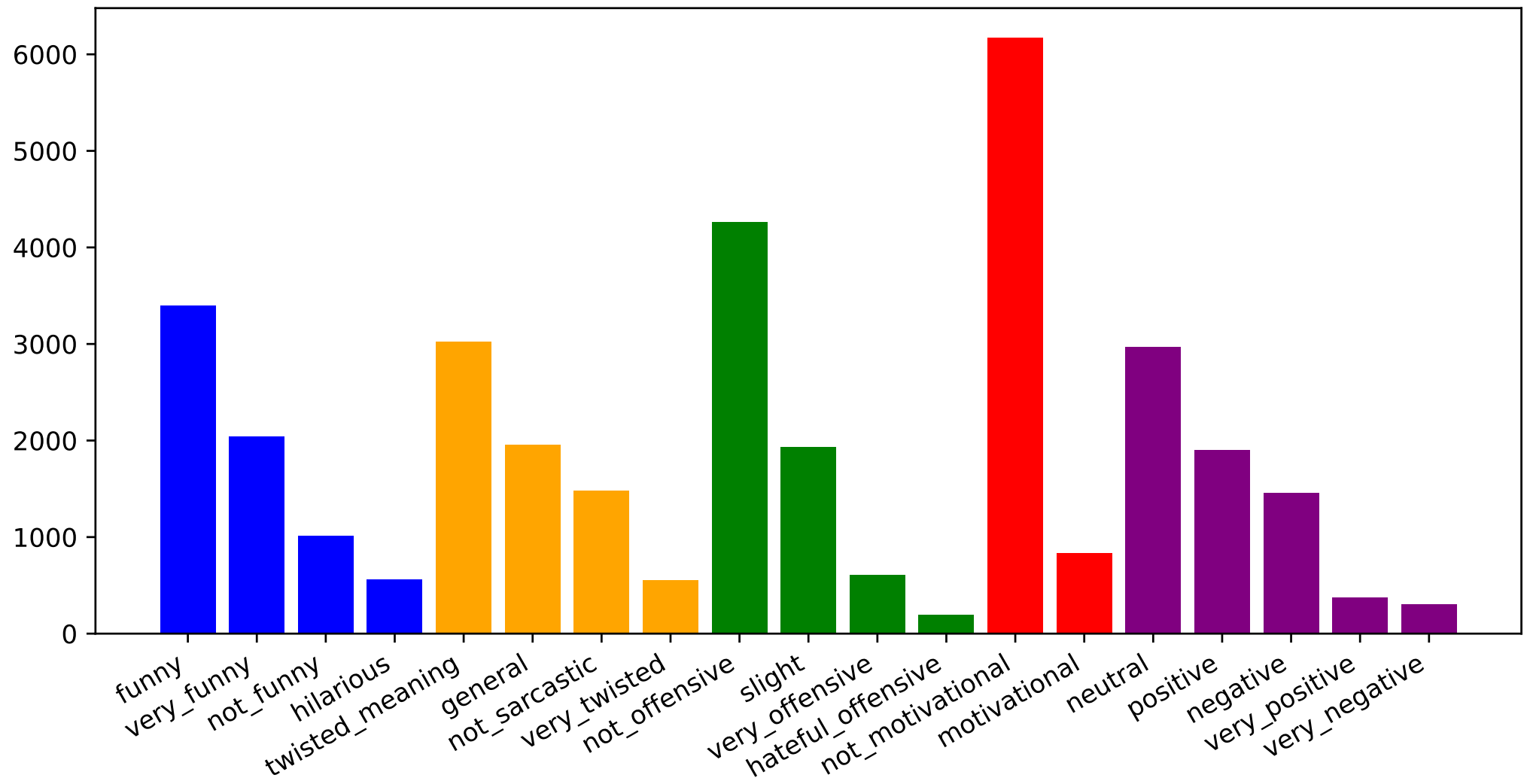

**Figure 6:** Distribution of the samples across all labels.

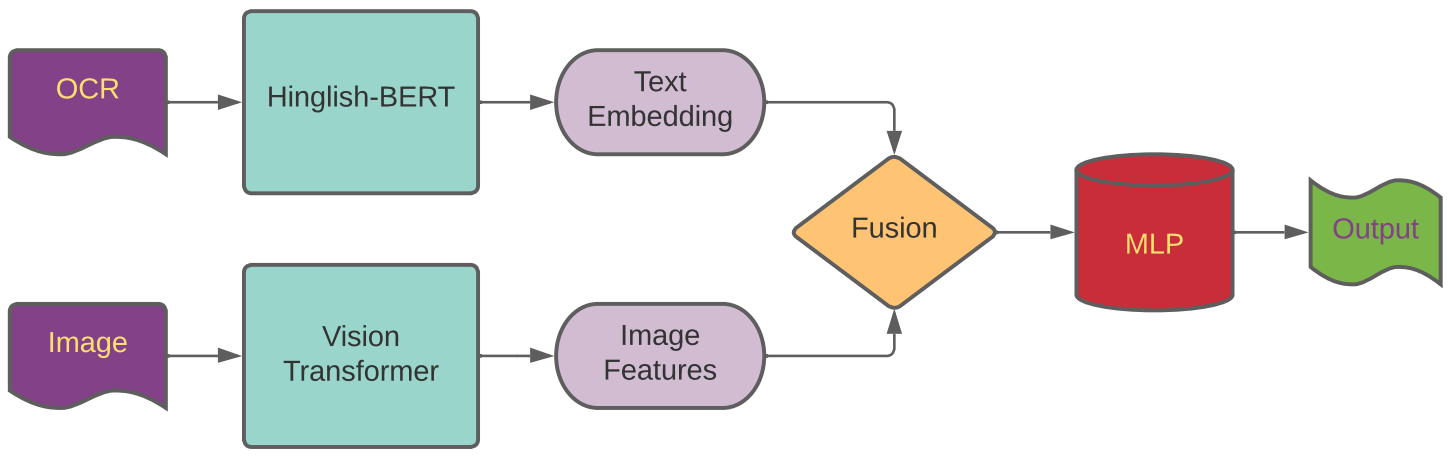

**Figure 7:** Baseline model architecture. It combines the text and image features for final classification. The hinglish BERT accounts for the code-mixing in data.

from the statistical features in Fig. 5, we can conclude that the emotions in memes overlap, demonstrating the difficulty of the tasks. A number of intriguing facts are revealed, including the fact that many offending memes are humorous. Additionally, a lot of the memes are humorous and lack inspiration, as can be seen. On average, the Code Mixed Index (CMI) [42] for the training, validation and test set is 14.94, 20.19 and 20.06 respectively.

## 5. Baseline model

The importance of considering both the visual and textual features is vital for multi-modal data, especially in the case of memes where the context can only be captured using a combination of both components. Attention models are exceptional at representing text with respect to context and a widely known model with strong performance is BERT [43]. As the dataset is not in English but instead in Hindi-English, we use a multilingual variant of BERT, specifically Hinglish-BERT from Verloop [44], with both the backbone and linear layers of the LM finetuned. The model is implemented using *BERT-base-multilingual-cased*, which is fine-tuned on Hinglish data. The visual features are obtained from the pre-trained Vision transformer model (ViT) [45]. The ViT model can outperform normal CNNs computationally and by accuracy, thanks to the positional embedding of image patches done by ViT. The pooled output from the ViT model is concatenated with the Hinglish-BERT embedding. The combined features are then classified after being passed through a MLP. With changes to the MLP, the multi-modal features are used for all three sub-tasks. The model architecture is displayed in Figure 7. The results for each task are provided in Table 1. The codes will be made available at https://github.com/Shreyashm16/Memotion-3.0.

## 6. Results

Baseline results in Table 1 show Weighted F1 scores for each task and sub-task. Using ViT for extracting visual features and Hinglish-BERT for the textual features, the baseline models scores 33.28% for Task A, 74.74% for Task B and 52.27% for Task C.

This dataset will be made public, and we leave it to future research to develop more sophisticated systems that go deeper into Memotion Analysis.

| Task | Class | Weighted F1 score |
|---|---|---|
| Task-A | Sentiment | 33.28% |
| Task-B | Humour | 84.55% |
| | Sarcasm | 74.82% |
| | Offensive | 48.84% |
| | Motivation | 90.78% |
| | Average | 74.74% |
| Task-C | Humour | 43.03% |
| | Sarcasm | 32.89% |
| | Offensive | 42.40% |
| | Motivation | 90.78% |
| | Average | 52.27% |

**Table 1**
Baseline scores (Weighted F1) of the baseline model on Memotion Analysis tasks.

## 7. Conclusion and Future Work

In this study, we present a hindi-english dataset for the challenge of sentiment in a multimodal environment. This is the first significant multimodal dataset for Hindi code-mixed meme categorization that we are aware of. We provide annotated data for three tasks, namely sentiment analysis, emotion classification, and strength of emotion, in order to provide a fine-grained and thorough analysis of memes. By combining the image features extracted from the ViT model and the textual features using Hinglish-BERT, and then passing these joint embeddings to a simple MLP, we design the baseline for the tasks. It should be mentioned that our models are preliminary and that more creative approaches will enhance performance much more. In the future, we intend to extend our work by designing a single model for all languages, instead of creating separate models for memes of different languages. We could also work on generating memes for the task, instead of collection to customize the dataset.